\documentclass[conference]{IEEEtran}

\usepackage{graphicx}
\usepackage{listings}
\usepackage{amsmath}
\usepackage{subcaption}
\usepackage[colorinlistoftodos]{todonotes}
\begin{document}
\title{General Video Game Rule Generation}

\author{\IEEEauthorblockN{Ahmed Khalifa}
\IEEEauthorblockA{Tandon School of Engineering\\
New York University\\
Brooklyn, New York 11201\\
Email: ahmed.khalifa@nyu.edu}
\and
\IEEEauthorblockN{Michael Cerny Green}
\IEEEauthorblockA{Tandon School of Engineering\\
New York University\\
Brooklyn, New York 11201\\
Email: mcg520@nyu.edu}
\and
\IEEEauthorblockN{Diego Perez-Liebana}
\IEEEauthorblockA{University of Essex\\
Colchester, United Kingdom\\
Email: dperez@essex.ac.uk}
\and
\IEEEauthorblockN{Julian Togelius}
\IEEEauthorblockA{Tandon School of Engineering\\
New York University\\
Brooklyn, New York 11201\\
Email: §julian@togelius.com}}


%


\maketitle
\begin{abstract}
We introduce the General Video Game Rule Generation problem, and the eponymous software framework which will be used in a new track of the General Video Game AI (GVGAI) competition. The problem is, given a game level as input, to generate the rules of a game that fits that level. This can be seen as the inverse of the General Video Game Level Generation problem. Conceptualizing these two problems as separate helps breaking the very hard problem of generating complete games into smaller, more manageable subproblems. The proposed framework builds on the GVGAI software and thus asks the rule generator for rules defined in the Video Game Description Language. We describe the API, and three different rule generators: a random, a constructive and a search-based generator. Early results indicate that the constructive generator generates playable and somewhat interesting game rules but has a limited expressive range, whereas the search-based generator generates remarkably diverse rulesets, but with an uneven quality.
\end{abstract}
\IEEEpeerreviewmaketitle
\section{Introduction}
A number of game-based AI competitions are run annually at academic conferences focused on AI and games, such as CIG. Most of these conferences are focused on playing games; competitors submit controllers that are evaluated on how well they play games. At the same time, procedural content generation (PCG) is an active area for Artificial Intelligence in Games research~\cite{shaker2016procedural}. One such competition is the Level Generation Track of the Mario AI Championship, where competitors submitted level generators which were judged based on their ability to generate levels for the game \cite{shaker20112010}; another example is the Level Generation Track of the General Video Game AI Competition, where competitors submit controllers that are tested on their ability to generate levels for any game within a particular domain~\cite{khalifa2016general}.

However, no competition has so far focused on generating \emph{rules} for video games, despite this being an active research topic. We believe that in order to make progress in video game rule generation, there must be a benchmark. This benchmark would allow researchers to compare their results and improve them by having access to previous work. We also believe it is important to start somewhat smaller than the full game generation problem, and focus on a well-delimited partial problem, in the light of the somewhat disappointing results of previous work attempting to generate levels and rules at the same time~\cite{barros2015towards}. In this paper, we introduce the General Video Game Rule Generation (GVG-RG) Framework. This framework is intended to be used as a benchmark for rule generation. Our framework uses VGDL to describe games, with the difference being that we will limit generation to only that of game rules and termination conditions. This paper introduces the GVG-RG track of the GVGAI Competition, as we believe competitions are the best way to encourage students, academics, and industry researchers to use the framework. 

This paper is structured as follows: Sections \ref{sec:background} and \ref{sec:vgdl} explains the required background to understand this paper, followed by section \ref{sec:gvgrg} explaining our framework (GVG-RG). Section \ref{sec:competition} explains the competition and how it will be organized. Section \ref{sec:methods} explains three sample rule generators provided with the framework, followed by the results from each generator in section \ref{sec:results}. Finally, section \ref{sec:conclusion} concludes the paper and talk about future works.

\section{Background}
\label{sec:background}

Many different types of game content can be procedurally generated, ranging from textures to quests to level to game rules. 
For example: SpeedTree~\cite{speedtree} is one of the most famous tools used to generate different trees and green areas with very high quality\footnote{http://www.speedtree.com/academy-award.php}, and widely used in commercial games. 
There are also numerous games that feature level generation, character generation, planet generation, etc. Overall, there are many types of content where PCG methods have proven effective and are in wide use.

This is not the case with rule generation. Generating rules requires defining a way to describe them. Game description languages are a descriptive languages that is used to define certain type/category of games. There are game description languages suited for defining board games~\cite{love2008general,browne2010evolutionary}, card games~\cite{font2013card}, videogames~\cite{ebner2013towards}, puzzle games~\cite{puzzlescript}, strategy games~\cite{mahlmann2011towards}, and so on. Having a more focused language helps the generator find good games. For example: using a description language for board games will help the generator to focus on finding good game rules instead of trying to understand the meaning of the game board.

Rule generation can be divided into two main categories: \emph{board games} and \emph{video games}. The earliest attempts were with \emph{Board games}, which is not surprising as board games have been around for thousands of years. One of the earliest attempts to tackle rule generation is METAGAME~\cite{pell1992metagame}. METAGAME is a research project by Pell to generate chess-like games. The generated games are symmetric, which means both players have the same pieces and use the same rules. 
The game generator does not use simulations but instead follows a certain grammar to generate playable games. 

Pell's system doesn't mind if the generated games are balanced or not. Balanced games are games that do not give advantage for a player over the other. Hom and Marks~\cite{hom2007automatic} improved over METAGAME by adding game simulations. Each generated game is played with AI against itself for a fixed amount of times, striving for a 50/50 win rate. 

Although the generated games are balanced from Hom and Marks' research, the search space is limited to chess-like games. Browne~\cite{browne2010evolutionary} introduced a simple description language that can describe different board games ranging from Tic-Tac-Toe to Go. Browne used an evolutionary technique to evolve new board games using a system called Ludi. Once the games were generated, Ludi used simulation to evaluate them. One of the generated games (Yavalath) is sold commercially and considered one of the top 100 best abstract games in 2011~\cite{yavalath}.


Video game rule generation has not been as successful as board game rule generation. A primary cause of this is the real-time aspect of video games compared to turn based games. Current research has not produced a general purpose AI agent capable of playing all games in the GVGAI game set~\cite{bontrager2016matching}. It is more difficult to use simulation-based evaluation of video games than board games for various reasons, including the discrepancy between optimal play and human-like play~\cite{khalifa2016modifying}.

Togelius and Schmidhuber~\cite{togelius2008experiment} generated arcade games, using 2D matrices to represent their interaction rules and score changes. The generated games were evaluated based on Koster's Theory of Fun~\cite{koster2013theory} where a good game is defined as one that the player can learn from. These generated arcade games were mostly simple chasing games due to the small search space. Smith and Mateas~\cite{smith2010variations,smith2010ludocore} did research on generating games as well, using Answer Set Programming (ASP) to define the search space. Instead of searching for new games, the game designer described the constraints required in the game, where a \textit{constraint solver} proceeded to carve the search space to only include games with these constrains. The generated games were similar to the games in Togelius and Schmidhuber work~\cite{togelius2008experiment}.

Cook's research revolves around automatic designing of video games using ANGELINA\footnote{http://www.gamesbyangelina.org/}, a computer system designed to generate different types of games. ANGELINA passed through many phases, starting as a simple arcade game generator~\cite{cook2011multi} like Togelius and Schmidhuber's work~\cite{togelius2008experiment}, and later able to generate full 3D games~\cite{cook2014ludus} for Ludumdare\footnote{Ludumdare is a 48-hours game jam where the competitors has to design and develop a game around a certain theme.}. 

Video Game Description Language~\cite{ebner2013towards,schaul2014extensible} (VGDL) is a language used to define a variety of games from Sokoban to Space Invaders. In 2015, Nielsen et al.~\cite{barros2015towards} used (VGDL) to generate new games, using evolutionary search where games were evaluated using relative algorithm performance~\cite{nielsen2015general}. Within the genetic engine, the initial population contained random generated games, human-designed games, and mutated versions of the two. The generated games are considered challenging but not as good as human-designed ones.


\section{Video Game Description Language}\label{sec:vgdl}
Video Game Description Language~\cite{ebner2013towards} (VGDL) is a language used to represent 2D arcade games (Pacman), action games (Space Invaders), and puzzle games (Sokoban). VGDL games consists of 2 main parts: game description and level description. The game description is responsible for holding information about game objects, behaviors, interactions and termination conditions. The game description consists of $4$ main parts:
\begin{itemize}
\item \textbf{Sprite Set:} a list of all game objects called game sprites. Each sprite has a type, which defines its behavior. For example, in Pacman a ghost is considered to be a \emph{RandomPathAltChaser} which means it chases normal pacman but flees from pacman after eating a power pellet.
\item \textbf{Interaction Set:} a list of all game interactions. Interactions only occur upon collision between two sprites. For example, In Pacman, if the player collides with a pellet, the latter will be destroyed and the score increases.
\item \textbf{Termination Set:} a list of conditions. These conditions define how to win or lose the game. These conditions can be dependent on sprites or on a countdown timer.
\item \textbf{Level Mapping:} a table of characters and sprite names. This table is used to decode the level description.
\end{itemize}
The level description contains a 2D matrix of characters where each character is decoded using the Level Mapping.

The General Video Game (GVG) framework is a Java implementation of VGDL. The framework was originally designed for the General Video Game AI (GVG-AI) competition~\cite{perez20162014}. GVG-AI competition is a general AI competition where competitors design AI agents that can play different unseen games efficiently. Each agent is provided with $40$ms to decide the next action. Multiple competition tracks were introduced later to the framework such as Level Generation~\cite{khalifa2016general}, Learning, and Multiplayer Planning~\cite{gainageneral}. In this paper, we will introduce the rule generation track (more details in Section \ref{sec:competition}).

\section{GVG-RG Framework}\label{sec:gvgrg}
The framework follows the same design philosophy of the level generation framework~\cite{khalifa2016general}. The framework allows the users to create their generators and test them against different VGDL games. 
The framework provides the generator with \emph{sprite level description (SLDescription)} object and in return, it expects the game interactions and termination conditions. The generator could provide the framework with an optional \emph{sprite set structure} hashmap object.

The SLDescription contains information about the current game sprites and the current level. This object provides the generator with functions to retrieve the game sprite information such as name, type, and related sprites. For example: the avatar in \emph{Zelda} game is named \emph{avatar}, is of type \emph{ShootAvatar} (it can move in all $4$ directions and shoots bullets),  and has \emph{sword} as related sprite (the avatar can shoot sword sprite). The \emph{SLDescription} object provides the generator with the current level in form of 2D string array. Each string is a comma separated list of the names of all the sprites in that location. In addition, the \emph{SLDescription} object 
can be used to simulate any game by providing it with interactions and termination conditions. 

The \emph{sprite set structure} hashmap maps between a string and an array of strings. The main advantage of providing the hashmap is to group multiple sprites under a single label. For example: In \emph{Space Invaders}, all aliens can be grouped under the label \emph{``Harmful''}. 

\section{Rule Generation Competition}\label{sec:competition}
The rule generation competition will run in the similar manner to the level generation competition~\cite{khalifa2016general}. The main difference instead of generating a level for a certain game, the aim is to generate a new game for a certain level. 
Instead of generating the whole game, the competitors are required only to generate the interaction set and the termination conditions while fixing the sprite set. This limitation decreases the generative space size, leading to generating better games.

The competitors will submit their code in form of zip file contain their generator code to the server. All rule generators are expected to return the interactions and the termination conditions within a fixed amount of time (in this paper, each generator was allowed 5 hours to generate one game) on a reference computer. The specification of the computer will be posted later on the competition website\footnote{http://www.gvgai.net/} (in this paper, we used a recent Mac book pro\footnote{2.9 GHz Intel Core i5 with 8 GB 1867 MHz DDR3}. If the generator takes more than the provided time, it will be disqualified and its results will be ignored. There is no restriction on the language used for the generator but we will only provide a Java interface.

The competitors will submit their generators code before the competition day to allow the system to generate the new games by the competition day. The generated games will be judged by humans on the competition day. The system will pick two random generated games and show them to the judges. Each judge will play each game as many times as required, then pick which game feels better to them. The system will ask the judge to briefly describe the picked game. All the results will be submitted to an SQL server and the most preferred generator will win the competition.

\section{Methods}\label{sec:methods}
We implemented three different sample rule generators to give the competitors a smooth start using the framework. Each of these generator uses a different technique to generate a new game. The following subsections describe these generators in detail.

\subsection{Random Rule Generator}
Random rule generator is the simplest of all. The main idea is to generate random interactions and termination conditions that the framework can run with no errors.

The generator follows these steps to generate game:
\begin{enumerate}
\item Pick a random number of interactions based on the number of sprites in the game.
\item Repeat the following steps till the required amount of interactions has been reached:
\begin{enumerate}
\item Pick two random sprites found in the current level including the end of screen (EOS).
\item Pick a random integer value for \emph{scoreChange}.
\item Pick a random interaction from all the available interactions. These interactions must compile with no errors with the current selected sprites. 
\end{enumerate}
\item After generating all the interactions, the system generates two termination conditions:
\begin{itemize}
\item \textbf{Winning Condition:} can be either a time out condition or a sprite counter condition. The time out condition fires after a randomly selected amount of time, while the sprite counter condition fires when the number of a randomly selected sprite (excluding the avatar) reaches $0$.
\item \textbf{Losing Condition:} fires when the avatar dies.
\end{itemize}
\end{enumerate}


\subsection{Constructive Rule Generator}
The constructive rule generator uses common game knowledge to generate games. For example: If there is an Non Playable Character (NPC) chasing another object, there is a high chance that the NPC will kill it upon collision. For example: Ghosts in \emph{Pacman} chase \emph{Pacman} to kill it. Using this knowledge, we can construct better games than random ones in a small amount of time.

Before generating the game, the constructive generator uses an instance of the \emph{LevelAnalyzer} class to understand the current level. The \emph{LevelAnalyzer} class is a helper class provided with the framework that calculates statistics of the game sprites based on the current level. It provides functions to get specific sprites based on their map coverage percentage and/or have a certain type. For example: We can get background sprites by asking the \emph{LevelAnalyzer} to get us \emph{Immovable} sprites with $100\%$ map coverage.

The constructive generator use the \emph{LevelAnalyzer} object to identify different sprite categories: 
\begin{itemize}
\item \textbf{Wall sprites:} are immovable sprites that surround the current level and cover at most $50\%$ of the entire level. If it doesn't exist, the end of screen (EOS) is considered the wall sprite.
\item \textbf{Score sprites and Spike sprites:} are immovable sprites that covers at most $10\%$ of the entire level.
\item \textbf{Avatar sprite:} is the sprite controlled by the player and exists in the current level.
\end{itemize}

The generator selects one sprite for each category then it follows the following steps to generate a new game:
\begin{enumerate}
\item \textbf{Get Resource Interactions:} \emph{collectResource} interactions are placed between \textit{avatar} and all sprites with type \emph{Resource}.
\item \textbf{Get Immovable Interactions:}  Interactions for \textit{score} sprite and \textit{spike} sprite added. Score sprite is collected by the avatar and gives the avatar $1$ score point. Spike sprite has $50\%$ chance to be either harmful or collectible. A harmful sprite kills the avatar upon collision, while a collectible sprite gives the avatar $2$ score points.
\item \textbf{Get NPCs Interactions:} The NPC interactions are different based on their type:
\begin{itemize}
\item \textit{Fleeing NPC:} is collected by the chaser sprite and adds $1$ point score.
\item \textit{Bomber NPC:} kills the avatar sprite upon collision. A Bomber NPC also spawns sprites. Spawned sprites have $50\%$ chance to be collected by the avatar for $1$ score point or kill the avatar upon collision.
\item \textit{Chaser NPC:} kills the chased sprite upon collision. If the chased sprite is not the avatar sprite, there is a $50\%$ chance to duplicate itself.
\item \textit{Random NPC:} has $50\%$ chance to kill the avatar or get killed by it for $1$ score point.
\end{itemize}
\item \textbf{Get Spawner Interactions:} The generator decides with $50\%$ either the spawned sprites will kill the avatar sprite upon collision or get collected by the avatar sprite for $1$ score point.
\item \textbf{Get Portal Interactions:} If a portal is of type \emph{``Door''}, the avatar can kill it upon collision. Otherwise, the avatar is teleported towards the portal destination.
\item \textbf{Get Movable Interactions:} The generator decides with $50\%$ chance if the moving sprites are collected by the avatar for $1$ point score or kills the avatar sprite.
\item \textbf{Get Wall Interactions:} The generator decides with $50\%$ if the walls should be normal walls or fire walls. Normal walls block all moving sprites from passing through them, while fire walls kill all moving sprites upon collision.
\item \textbf{Get Avatar Interactions:} This step only happens if the avatar can shoot bullets. The generator adds interactions between harmful sprites and the avatar's bullets. Harmful sprites are any sprite that can kill the avatar. Harmful sprites are killed upon collision with avatar bullets.
\item \textbf{Get Termination Conditions:} The generator adds two termination conditions:
\begin{itemize}
\item \textit{Winning Condition:} The generator chooses an applicable condition from the following list: the avatar reaches a door sprite, all harmful sprites are dead, all fleeing NPCs are dead, all collectible sprite are dead (collectible sprites are immovable sprites that add score points), or time runs out.
\item \textit{Losing Condition:} if avatar dies, the game is lost.
\end{itemize}
\end{enumerate}

\subsection{Search Based Rule Generator}
The search based rule generator uses Feasible-Infeasible 2 Population Genetic Algorithm (FI2Pop)~\cite{kimbrough2008feasible} to search for new games. FI2Pop is a genetic algorithm which keeps track of two populations (Feasible and Infeasible population). The infeasible population tries to satisfy multiple constraints, while the feasible population tries to improve the overall fitness. At any time, if a chromosome failed to satisfy the constraints it is moved to the infeasible population, and if any chromosome satisfies all constrains, it will be moved to the feasible population.

Before evolution begins, an initial population of $50$ chromosomes is created from a mix of games generated by the random rule generator and the constructive rule generator. Based on preliminary experiments, The population consists of a mix of $40\%$ random rules and $20\%$ constructive rules. The rest of the population ($40\%$) is created by mutating over the constructive generated games. Mutation will be discussed in depth later in this section. After initialization, each chromosome undergoes \emph{cleansing} where duplicated rules are stripped out of the ruleset. The newly created population is then evaluated for fitness. A chromosome is \textit{feasible} if it tests positively on three conditions:
\begin{itemize}
\item The chromosome's ruleset does not generate any \textit{errors} in the GVG-AI engine. Errors are classified as anything that prevents the engine from simulation.
\item A do-nothing agent which makes no moves will survive the first $40$ steps (1.6 seconds) of playtime.
\item The number of \textit{bad frames} simulated during random, agent, and smart agent is calculated. \textit{Bad frames} are defined as frames in which game sprites are drawn outside the boundaries of the screen. If $>$ 30\% of the frames are bad frames, then this chromosome is infeasible.
\end{itemize}

The feasibility of the chromosome is calculated as:
\begin{equation}
f_{c} = 0.3 * \frac{1}{n_{e} + 1} + 0.2 * \frac{n_{dna}}{40} + 0.2 * \frac{1}{n_{w} + 1}+ 0.3 * (1-\frac{n_{bf}}{n_{f}})
\end{equation}
where $n_{e}$  and $n_w$ are the number of errors and warnings generated by the GVG-AI engine respectively, $n_{dna}$ is the number of frames that do-nothing survives, $n_{bf}$ is the number of bad frames found during the play-throughs, and $n_{f}$ is the maximum number of frames allowed in this game (used for normalization).

If a chromosome passes the feasibility test, then its true fitness is evaluated. This is done using relative algorithm performance~\cite{nielsen2015general}. We use three agents: a \textit{smart} agent which uses Open Loop MCTS (OLETS), the winner of the GVG-AI competition in 2014~\cite{perez20162014}, a \textit{baseline} agent which uses Monte Carlo Tree Search (MCTS), and a \textit{random} agent which randomly selects its next move. Each agent is allowed 3 independent play-throughs, with the results averaged.

The smart agent plays first and its step count is recorded for the best game it plays. The baseline and the random agent then play the game for as many steps as the smart agent.
The three agents' scores and win rates are averaged across their play-throughs. A score value is then calculated for each agent according to the following equation:
\begin{equation}
O = 0.9 * A_{win} + 0.1 * Sigmoid(x)
\label{eq:score}
\end{equation}
where $A_{win}$ is the averaged win count and $x$ is the average game score. To calculate the final score value, we take the difference between the smart and the baseline agent, and multiply it with the difference between the baseline and the random agent:
\begin{equation}
S_{final} = (O_{smart} - O_{baseline}) * (O_{baseline} - O_{random})
\end{equation}

Throughout all play-throughs, we keep track of the number of distinct triggered interactions. After the play-throughs are complete, we divide it by the number of rules in the ruleset.
\begin{equation}
S_{rules} = \frac{R_{unique}}{R_{total}}
\end{equation}
where $R_{unique}$ is the number of unique interactions fired during the play-throughs and $R_{total}$ is the total number of interactions in the current chromosome. The average game length over all play-throughs is tracked. This is done to penalize games that are shorter than 500 frames (20 seconds), as these games are not long enough to be considered playable. The overall fitness is calculated in the following formula:
\begin{equation}
f_c = S_{final} * S_{rules} * S_{gameLength}
\end{equation}

After being evaluated for fitness, the chromosomes under-go rank selection inside their respective populations in order to create the next generation. If selected, two children are created from clones of the parents. The children then have a $90\%$ chance to undergo \textit{crossover}. If crossover does not occur, these children then have a chance undergo mutation. If not selected for mutation, these children are simply cloned into the new population. We use one point crossover between chromosomes where a random section of each child's ruleset is swapped.

\textit{Mutation} for interaction and termination sets is more complicated than crossover. There are 3 types of mutation, and each type contains two sub-types. 
\begin{itemize}
\item \textit{Insertion} is divided into insertion of a new parameter into an existing rule, and insertion of an entirely new rule.
\item \textit{Deletion} is divided into deletion of a parameter from an existing rule, and deletion of an entire rule.
\item \textit{Modify} is divided into modification of an existing rule parameter, and modification of the rule itself.
\end{itemize}

Chromosomes have a 10\% chance to be mutated upon, and undergo up to two mutations. Within a single round of mutation, a chromosome has a 50\% chance to mutate on their interaction set, and a 50\% chance to otherwise mutate on their termination set. 2\% elitism (1 chromosome) was used to preserve the best games.  After timeout, the ruleset of the chromosome with the highest fitness is returned as the generated ruleset.

\section{Results}\label{sec:results}
\begin{figure}[!t]
\centering
\includegraphics[width=0.9\linewidth]{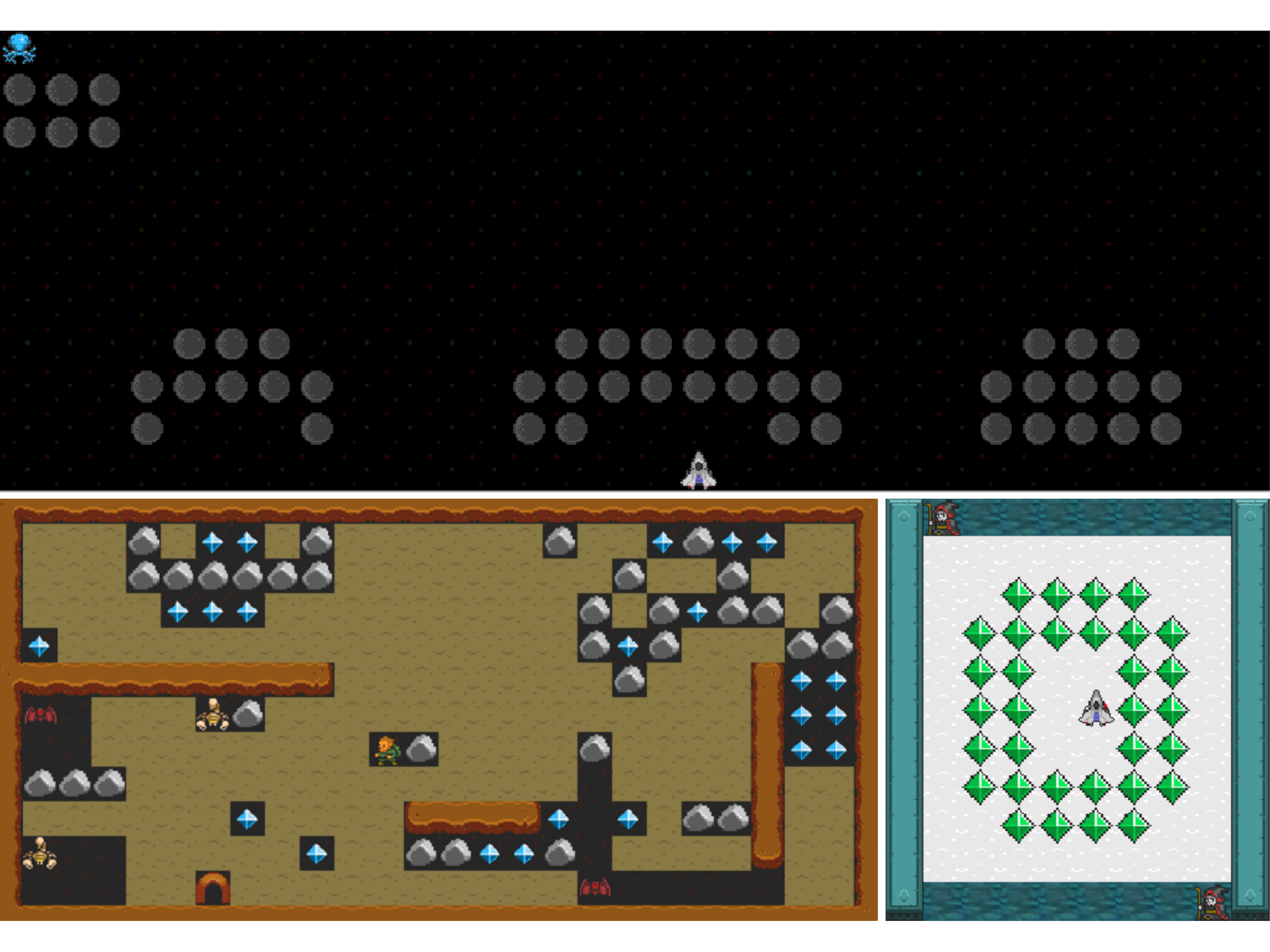}
\caption{The VGDL experimental levels in order from top to bottom right: Aliens, Boulderdash, and Solarfox.}
\label{fig:gvgaigames}
\end{figure}
We tested our generators against 3 different games:
\begin{itemize}
\item \textbf{Aliens:} is a port of Space Invaders. The goal is to shoot all the aliens before killing the player.
\item \textbf{Boulderdash:} is a port of Boulderdash. The goal is to collect at least 10 diamonds before heading towards the exit while avoiding enemies.
\item \textbf{Solarfox:} is a port of Solarfox. The goal is to collect all the diamonds without hitting walls or enemy bullets.
\end{itemize}
These games are selected because they vary in their sprite set and their level design. For example: The avatar is \emph{FlakAvatar} in Aliens, \emph{ShootAvatar} in Boulderdash, and \emph{OngoingAvatar} in Solarfox. Figure \ref{fig:gvgaigames} shows the levels used for each of the above games. The levels are distinguishable from each other. For example: Aliens have a very few sprites in the level compared to Boulderdash, while Solarfox has a very small level dimensions. The following subsections shows some of generated games using our generators.

\begin{figure*}
\begin{subfigure}{.33\textwidth}
  \centering
  \includegraphics[width=\linewidth]{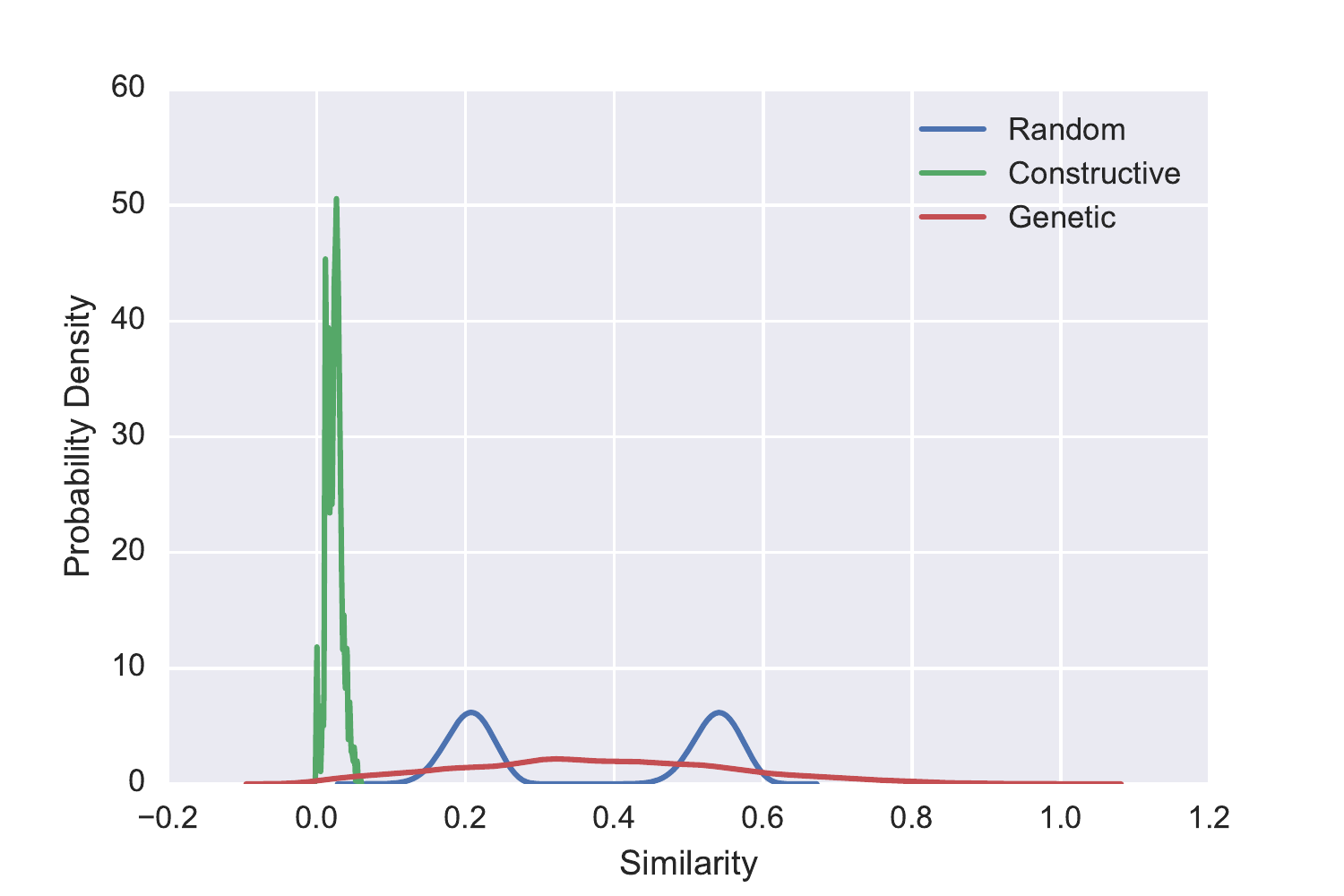}
  \caption{Aliens}
  \label{fig:aliensSim}
\end{subfigure}%
\begin{subfigure}{.33\textwidth}
  \centering
  \includegraphics[width=\linewidth]{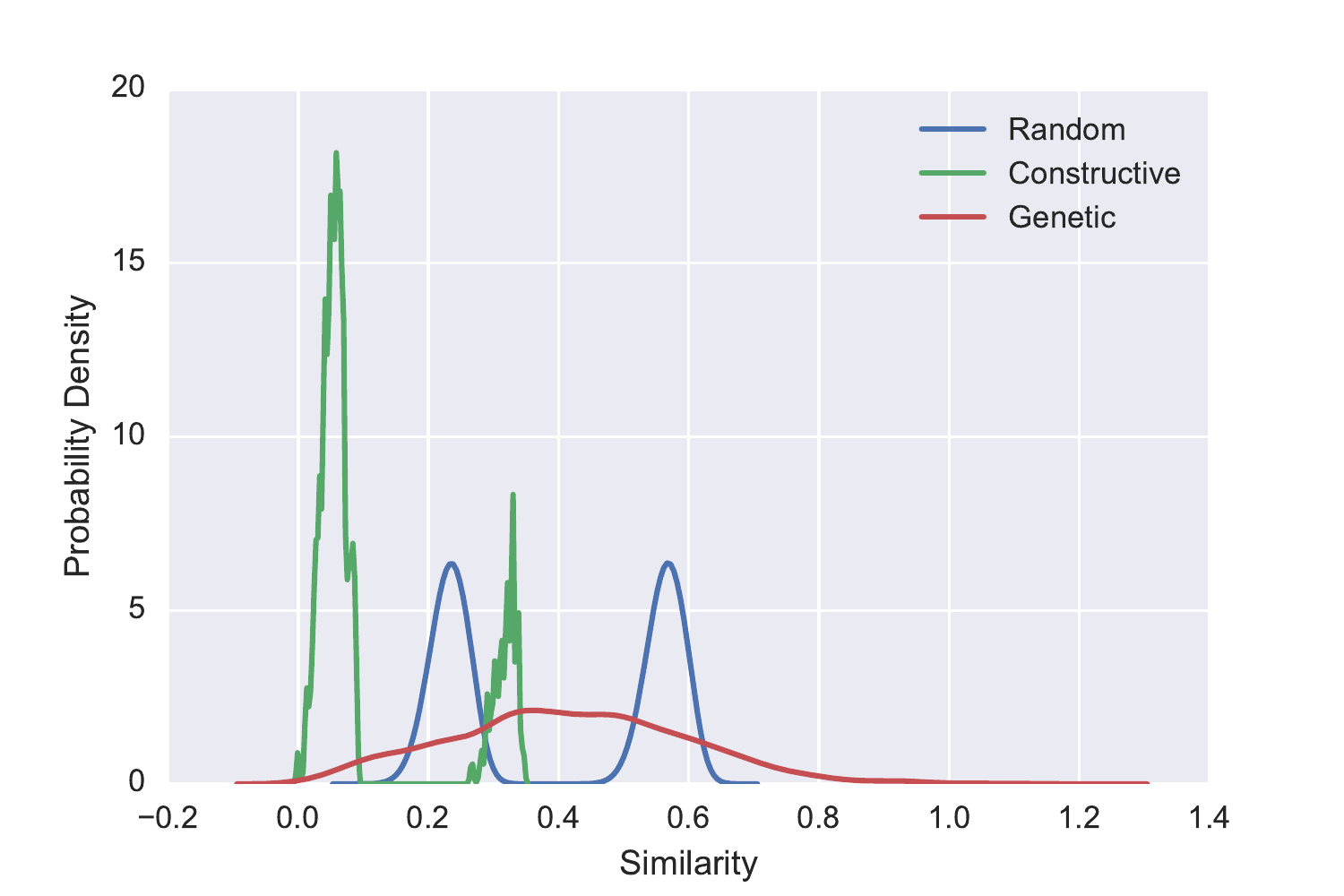}
  \caption{Boulderdash}
  \label{fig:boulderdashSim}
\end{subfigure}
\begin{subfigure}{.33\textwidth}
  \centering
  \includegraphics[width=\linewidth]{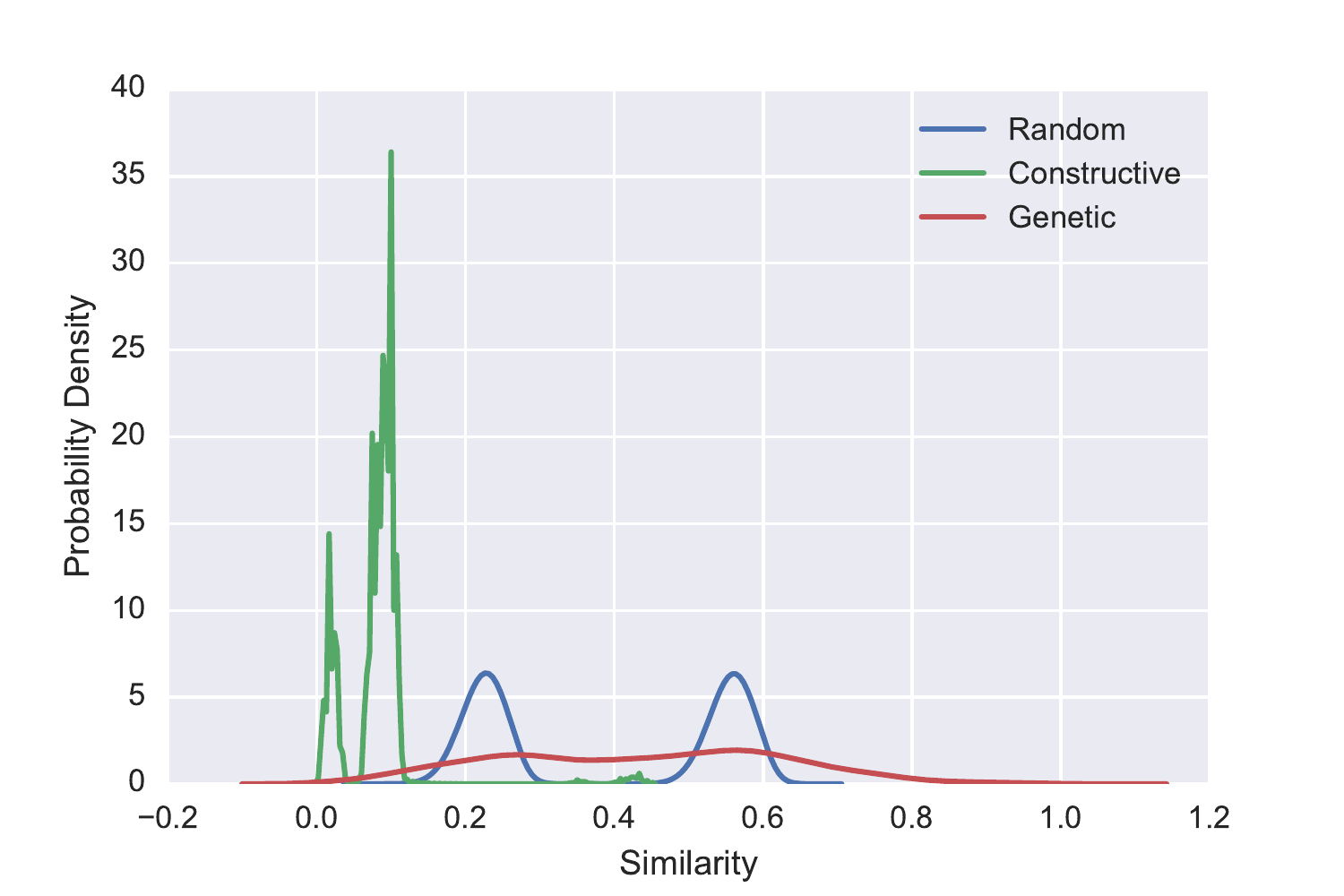}
  \caption{Solarfox}
  \label{fig:solarfoxSim}
\end{subfigure}
\caption{Probability density of the similarity metric between the generated games. ``$0$'' means the games are identical, while ``$1$'' means the games are totally different.}
\label{fig:simlarity}
\end{figure*}

Figure \ref{fig:simlarity} shows the distribution of the similarity between the generated games by all generators. There are 1000 constructive and random games, and 350 search-based games (due to time constraints). The similarity is calculated by comparing the distance between interactions and termination conditions from one game with all the other games and picking the minimum distance. All comparisons are done in the same manner: for a given interaction or termination condition, all parts will be compared to all other parts of other interaction or termination conditions in every other game. Parts that are identical are given a score of $0$, and parts that are not are given a score of $1$. 
At the end of this process, the game will have a score that will be scaled to be between 0 and 1. The constructive generator has the most similar games compared to the random or the search based generators, while the search based generator covers the highest area. The random generator covers a small portion of space due to the limit of the number of the generated interactions which is discussed in the next subsection.


\subsection{Random Game Generation}
This section shows the three generated games using the \emph{Random Rule Generator}. All interactions and termination conditions are selected randomly. There can only be a max of  $5$ interactions to minimize the number of useless interactions.

Listing \ref{ex:randomAliens} shows the generated interactions and termination conditions for Aliens. The only way to win the game is to wait till the timer is out. If the avatar tries to go outside the screen bounds, all sprites (including the avatar) will freeze till timeout. The player wins the game by waiting for 1200 frames.

\begin{lstlisting}[language=python,basicstyle=\tiny,caption={A randomly generated Aliens interaction set and termination conditions.},label={ex:randomAliens},frame=single,tabsize=4,captionpos=b,breaklines=true,breakatwhitespace=false]
TerminationSet
	Timeout limit=1200 win=True
	SpriteCounter stype=avatar limit=0 win=False
InteractionSet
	EOS base > stepBack scoreChange=1
	avatar EOS > turnAround
	base portalSlow > flipDirection scoreChange=-2
	EOS avatar > flipDirection scoreChange=-1
	EOS avatar > undoAll scoreChange=1
\end{lstlisting}

Listing \ref{ex:randomBoulderdash} shows the generated interactions and termination conditions for Boulderdash. None of the interactions ever trigger except for the first one where it adds health point to bats upon collision with any wall. The player wins the game by waiting for 800 frames.

\begin{lstlisting}[language=python,basicstyle=\tiny,caption={A randomly generated Boulderdash interaction set and termination conditions.},label={ex:randomBoulderdash},frame=single,tabsize=4,captionpos=b,breaklines=true,breakatwhitespace=false]
TerminationSet
	Timeout limit=800 win=True
    SpriteCounter stype=avatar limit=0 win=False
InteractionSet
	bat wall > addHealthPointsToMax
    diamond exitdoor > addHealthPoints scoreChange=0
    background wall > undoAll
    wall dirt > undoAll scoreChange=-2
    scorpion background > attractGaze
\end{lstlisting}

Listing \ref{ex:randomSolarfox} shows the generated interactions and termination conditions for Solarfox. The only way to get points is by colliding with the top enemy. The avatar needs to catch the top enemy as fast as possible before getting outside the game level. The player wins the game after 1000 frames.

\begin{lstlisting}[language=python,basicstyle=\tiny,caption={A randomly generated Solarfox interaction set and termination conditions.},label={ex:randomSolarfox},frame=single,tabsize=4,captionpos=b,breaklines=true,breakatwhitespace=false]
TerminationSet
	Timeout limit=1000 win=True
	SpriteCounter stype=avatar limit=0 win=False
InteractionSet
	wall diamond > teleportToExit scoreChange=1
	top avatar > cloneSprite scoreChange=1
	wall diamond > bounceForward scoreChange=2
	enemyground EOS > removeScore
	avatar diamond > pullWithIt
\end{lstlisting}

\subsection{Constructive Game Generation}
This section shows the generated games by the \emph{Constructive Rule Generator}. All the generated games by the constructive are highly similar due to the predefined interactions for each sprite category. 

Listing \ref{ex:constAliens} shows the generated interactions and termination conditions for Aliens. The generated game is similar to the original game with two main differences. The first difference is all bullets reflect when they reach EOS. The second difference is enemies never go down towards the avatar. The player wins when all the aliens die.

\begin{lstlisting}[language=python,basicstyle=\tiny,caption={A constructive generated Aliens interaction set and termination conditions.},label={ex:constAliens},frame=single,tabsize=4,captionpos=b,breaklines=true,breakatwhitespace=false]
TerminationSet
	SpriteCounter stype=harmful limit=0 win=True
	SpriteCounter stype=avatar limit=0 win=False
InteractionSet
	portalSlow avatar > killSprite scoreChange=1
	avatar alienGreen > killSprite
	bomb avatar > killSprite scoreChange=1
	avatar alienBlue > killSprite
	avatar bomb > killSprite
	avatar alienBlue > killSprite
	avatar bomb > killSprite
	avatar EOS > stepBack
	sam EOS > flipDirection
	bomb EOS > flipDirection
	alienGreen EOS > turnAround
	alienBlue EOS > turnAround
	alienGreen sam > killSprite scoreChange=1
	sam alienGreen > killSprite
	alienBlue sam > killSprite scoreChange=1
	sam alienBlue > killSprite
	bomb sam > killSprite scoreChange=1
	sam bomb > killSprite
	alienBlue sam > killSprite scoreChange=1
	sam alienBlue > killSprite
	bomb sam > killSprite scoreChange=1
	sam bomb > killSprite
\end{lstlisting}

Listing \ref{ex:constBoulderdash} shows the generated interactions and termination conditions for Boulderdash. The goal is to reach the exit door while avoiding chaser enemies (scorpions). The player can get points by collecting bats or catching the falling boulders before they get out of the screen. 

\begin{lstlisting}[language=python,basicstyle=\tiny,caption={A constructive generated Boulderdash interaction set and termination conditions.},label={ex:constBoulderdash},frame=single,tabsize=4,captionpos=b,breaklines=true,breakatwhitespace=false]
TerminationSet
	SpriteCounter stype=exitdoor limit=0 win=True
	SpriteCounter stype=avatar limit=0 win=False
InteractionSet
	diamond avatar > collectResource
	avatar scorpion > killSprite
	bat avatar > killSprite scoreChange=1
	exitdoor avatar > killSprite
	boulder avatar > killSprite scoreChange=1
	avatar wall > stepBack
	pickaxe wall > turnAround
	boulder wall > turnAround
	scorpion wall > wrapAround
	bat wall > wrapAround
	scorpion pickaxe > killSprite scoreChange=1
	pickaxe scorpion > killSprite
\end{lstlisting}

Listing \ref{ex:constSolarfox} shows the generated interactions and termination conditions for Solarfox. The new game is an easier version of the original game. The avatar needs to collect all the diamonds while avoiding being hit by any of the enemies (the top enemy or the bottom enemy). The avatar wins after 700 time steps.

\begin{lstlisting}[language=python,basicstyle=\tiny,caption={A constructive generated Solarfox interaction set and termination conditions.},label={ex:constSolarfox},frame=single,tabsize=4,captionpos=b,breaklines=true,breakatwhitespace=false]
TerminationSet
	Timeout limit=700 win=True
	SpriteCounter stype=avatar limit=0 win=False
InteractionSet
	diamond avatar > killSprite scoreChange=1
	avatar top > killSprite
	avatar bottom > killSprite
	avatar upshot > killSprite
	downshot avatar > killSprite scoreChange=1
	avatar EOS > stepBack
	diamond EOS > turnAround
	top EOS > turnAround
	bottom EOS > turnAround
	upshot EOS > turnAround
	downshot EOS > turnAround
\end{lstlisting}


\subsection{Search-Based Game Generation}
This section shows the generated games by the \emph{Search Based Rule Generator}. Overall, the generated games are more diverse than \emph{Constructive Generator}.

Listing \ref{ex:searchAlien} shows the generated interactions and termination conditions for Aliens. In this game, the avatar is always losing points as long as the avatar is in the level. The avatar stops losing points and its score is reseted to $0$ upon collision with EOS. The avatar wins after 1084 steps.

\begin{lstlisting}[language=python,basicstyle=\tiny,caption={A search based generated Aliens interaction set and termination conditions.},label={ex:searchAlien},frame=single,tabsize=4,captionpos=b,breaklines=true,breakatwhitespace=false]
TerminationSet
	Timeout limit=1084 win=True
	SpriteCounter stype=avatar limit=0 win=False
InteractionSet
	background base > align
	background avatar > reverseDirection scoreChange=-1
	EOS avatar > removeScore
	avatar background > reverseDirection scoreChange=-2
	base portalSlow > undoAll scoreChange=-1
\end{lstlisting}

Listing \ref{ex:searchBoulderdash} shows the generated interactions and termination conditions for Boulderdash. The generated game is a score based game where the avatar tries to get the highest score before going for the exit door. The avatar scores by collecting boulders, diamonds, and bats. The avatar must avoid colliding with the chaser enemies (scorpion). The avatar gets points when it kills the scorpions using the pickaxe. 

\begin{lstlisting}[language=python,basicstyle=\tiny,caption={A search based generated Boulderdash interaction set and termination conditions.},label={ex:searchBoulderdash},frame=single,tabsize=4,captionpos=b,breaklines=true,breakatwhitespace=false]
TerminationSet
	SpriteCounter stype=exitdoor limit=0 win=True
	SpriteCounter stype=avatar limit=0 win=False
InteractionSet
	avatar wall > stepBack
	exitdoor avatar > killSprite
	boulder avatar > killSprite scoreChange=1
	bat wall > flipDirection
	boulder wall > flipDirection
	scorpion wall > flipDirection
	pickaxe wall > flipDirection forceOrientation=avatar
	bat avatar > killSprite scoreChange=1
	scorpion pickaxe > killSprite scoreChange=1
	diamond avatar > collectResource
	avatar scorpion > killSprite
	pickaxe scorpion > killSprite
\end{lstlisting} 

Listing \ref{ex:searchSolarfox} shows the generated interactions and termination conditions for Solarfox. In this game, the avatar scores by collecting diamonds and colliding with the top and bottom enemies while avoid getting hit by their fireballs (\emph{upshot} and \emph{downshot}). The avatar wins after 800 frames.

\begin{lstlisting}[language=python,basicstyle=\tiny,caption={A search based generated Solarfox interaction set and termination conditions.},label={ex:searchSolarfox},frame=single,tabsize=4,captionpos=b,breaklines=true,breakatwhitespace=false]
TerminationSet
	SpriteCounter stype=avatar limit=0 win=False
	Timeout limit=800 win=True
InteractionSet
	top EOS > wrapAround
	avatar downshot > killSprite
	upshot EOS > wrapAround
	top avatar > killSprite scoreChange=1
	avatar EOS > stepBack
	diamond avatar > killSprite scoreChange=1
	bottom EOS > wrapAround
	diamond EOS > wrapAround
	avatar upshot > killSprite
	bottom avatar > killSprite scoreChange=1
	downshot EOS > wrapAround
\end{lstlisting}



\subsection{User Study}
\begin{table}[!t]
\centering
\begin{tabular}{|l|ccc|}
\hline
 & Aliens & Boulderdash & Solarfox\\
\hline
Search vs Rnd & 2/8 & 7/7 & 11/15\\
Search vs Const & 0/14 & 8/14 & 6/18\\
Const vs Rnd & 9/10 & 10/11 & 4/5\\
\hline
\end{tabular}
\newline
\caption{Comparison between our rule generators where ``Search'' is the search-based generator, ``Rnd'' is random generator, or ``Const'' is constructive generator. The first value is the number of times the user preferred the first game over the second. The second value is the total number of comparisons.}
\label{tab:userstudy}
\end{table}

To verify our results we ran a user study. In this study, the user is subjected to two generated games with the same level. These games are selected randomly from any generator. The user has to play each game for any number of times and select which game is preferable. The user can choose any of the following answers: ``First game is better'', ``Second game is better'', ``Both are good games'', or ``Neither of them is good''.

We collected 161 comparisons. Table \ref{tab:userstudy} shows the results for the user study by taking only ``First game is better'' and ``Second game is better'' in consideration. The result shows that all generators are better than the random generator in all the games except for ``Aliens''. In ``Aliens'', the random is only better than the search-based generator. By analyzing the generated games from the search-based generator, we found that all the game sprites are frozen (can't move) until the game times out. This happens due to the ``Bad Frames'' constraint where using a ``undoAll'' interaction (freezes all sprites) is the easiest way to satisfy it.

Looking at the constructive generator results, we can see that it beats the search-based in 2 games out of 3 (``Aliens'' and ``Solarfox''). The constructive is better in ``Aliens'' for the same reason stated before, but ``Solarfox'' was not expected as the search based games are as good as the constructive games. By looking at the generated games from both generators, we can see that the search based has one game where all sprites are frozen and can not move. We assume that it is the main reason behind the bias toward the constructive generator. 

\section{Conclusion \& Future work}\label{sec:conclusion}
This paper discussed the General Video Game Rule Generation problem, and introduced a software framework for working on this problem which will also be used as part of a new track of the GVGAI competition. As the framework is built on top of the GVGAI software, it benefits from all of the games, levels, agents, tools and APIs that have been developed as part of the overall GVGAI meta-project. The usefulness of this was clearly demonstrated in the experiments in this paper, which built on levels made for three of the roughly $100$ games that have been developed in the GVGAI framework, and used agents developed as part of the GVGAI competition for simulation-based testing of the rulesets as part of the search-based rule generator.

The main contributions of this paper are the formulation of the problem and presentation of the framework. While the experiments were mostly made to test the framework and gauge the difficulty of the problem, they are worth discussing in their own right. The constructive generator, which follows a simple step-by-step recipe, turns out to be able to produce playable games of some novelty, but with very limited diversity. The search-based generator displays a much broader expressive range, but the fitness function does apparently not guarantee neither playability nor interestingness.

The most obvious future research direction is to create rule generators that generate better rules. It stands to reason that this will eventually be possible within a search-based framework; this is a long-term project of ours, where in previous work the problems observed in trying to evolve VGDL rules and levels together~\cite{nielsen2015general} inspired the development of more human-like GVGAI-playing agents~\cite{khalifa2016modifying}, and ultimately this paper.
We hope you will help with this. The definition of the framework and the new competition track is meant to spur broader participation among the research community.

Once there is a method for generating good rulesets for individual levels within this restricted problem, we can start easing the restrictions: including more levels as inputs, allowing the generator to generate the sprite set and the level as well, or perhaps numerical parameters of the game engine. Another route is to not broaden the scope of the generator, but rather run different generators in sequence, such as running level generators on rule generator outputs and vice versa.
In either case, we believe the modularization of the larger game generation problem into multiple smaller generative problems with well-defined interfaces allows for many exciting possibilities for moving forward on the game generation problem.




%
\bibliographystyle{IEEEtran}
\bibliography{IEEEexample}

\end{document}